\documentclass{article}
\usepackage[preprint]{colm2026_conference}
\usepackage{times}
\usepackage{microtype}
\usepackage{amsmath,amssymb}
\usepackage{booktabs}
\usepackage{graphicx}
\usepackage{url}
\usepackage{lineno}
\usepackage{hyperref}
\definecolor{darkblue}{rgb}{0, 0, 0.5}
\hypersetup{colorlinks=true, citecolor=darkblue, linkcolor=darkblue, urlcolor=darkblue}

\title{The Fragility of Reasoning-Importance Attribution\\ for Pruning Language Models}

\author{Himanshu Mishra \\
University of British Columbia \\
\texttt{himmishra@student.ubc.ca}}

\newcommand{\ta}{\ensuremath{T_{\mathrm{ans}}}}
\newcommand{\tc}{\ensuremath{T_{\mathrm{cot}}}}
\newcommand{\tr}{\ensuremath{T_{\mathrm{rec}}}}

\begin{document}

\ifcolmsubmission
\linenumbers
\fi

\maketitle

\begin{abstract}
Attribution-guided pruning scores components by their causal contribution and protects the
important ones, and it is increasingly applied to compress reasoning language models without
destroying their chain of thought. Its reliability rests on an unexamined assumption: that
reasoning-importance can be faithfully measured. Across four small reasoning models we show this
measurement is fragile in three ways, each of which changes pruning decisions.
First, scoring final-answer tokens, the natural choice, saturates in \emph{every}
chain-of-thought model we test (answer-token negative log-likelihood, NLL, of $10^{-4}$--$10^{-3}$) because the model states
its answer inside its own reasoning; guiding head pruning with this target costs $27.6$ and
$32.2$ accuracy points relative to scoring reasoning tokens on the two models we power with five
seeds, while perplexity barely moves and even reverses sign at 7B. Second, causal importance is
not localizable to single units: single-neuron effects sit at the measurement floor, their
cross-split ranking is noise, and they cancel: a group's ablation effect disagrees in sign with
the summed effects of its members in 90\% of groups. Reproducible importance emerges only at
groups of $\sim\!100$ neurons. Third, contrastively isolated ``reasoning-specific'' importance is
not separable from generic importance, and protecting it is worse than protecting raw importance.
We additionally pre-registered the hypothesis that distillation-induced concentration of
importance predicts the pruning cost; it is not supported, and since our grid confounds
concentration with scale we report it as unresolved rather than reversed. The result is a short
set of measurement rules: score reasoning tokens, attribute at coarse granularity, and evaluate
with accuracy rather than perplexity.
\end{abstract}

\section{Introduction}\label{sec:intro}
Small language models with chain-of-thought (CoT) reasoning \citep{wei2022chain} are attractive
targets for on-device deployment, and pruning is the standard route to shrink them. Classical
pruning scores parameters by their contribution to language-modeling loss
\citep{han2015learning,frantar2023sparsegpt,sun2024wanda}, and a recent line of work makes
pruning \emph{reasoning-aware}: calibrating on self-generated CoT activations
\citep{lucas2025rac} or attributing importance causally and protecting high-scoring components
\citep{sheth2026cap}, echoing interpretability findings that specific heads carry outsized
functional load \citep{michel2019sixteen,voita2019analyzing}.

All such methods share an unexamined dependency: they assume the reasoning-importance signal is
faithful. Attribution requires three implicit choices: \emph{what} to score against (the target
token set), \emph{at what granularity} to score, and \emph{whether} reasoning-specific components
are separable from generically important ones. We show each choice is fragile in reasoning
models, and that the first has large, measurable consequences for pruning.

Our thesis is restatable in one sentence: \emph{to guide pruning without silently degrading
reasoning, importance must be scored on reasoning tokens rather than answer tokens, measured over
groups of at least $\sim\!100$ units rather than single components, and taken as raw importance
rather than a reasoning-specific contrast.} We frame this
as a methodological complement to reasoning-aware pruning, not a critique of any method. We
verified from source that the two closest methods do not use the saturating target: the
reconstruction-based pruner of \citet{lucas2025rac} reconstructs weights over prompt and decode
tokens, which include the chain of thought, and the causal-attribution pruner of
\citet{sheth2026cap} scores a span that is not the model's own CoT followed by its answer, and is
evaluated on non-distilled instruct models. The saturating target is a \emph{naive-but-plausible} choice
that we characterize, not a flaw we attribute to prior work.

\paragraph{Contributions.}
\begin{itemize}\itemsep2pt
\item \textbf{Target fragility.} Final-answer-token attribution saturates in all four models
(Section~\ref{sec:saturation}) and, used for head pruning, costs $+27.6$ (D1) and $+32.2$ (N1)
accuracy points versus scoring CoT tokens, with tight hierarchical-bootstrap intervals over five
seeds (Section~\ref{sec:target}).
\item \textbf{Granularity fragility.} Single-neuron causal importance is ill-posed: effects are
irreproducible and cancel (sign disagreement $0.90$); reproducible importance emerges only at
$G^*\!\approx\!100$ (Section~\ref{sec:granularity}).
\item \textbf{Specificity fragility.} Reasoning-specific importance collapses to its own noise
floor, and protecting the specific set is worse than protecting the generic set at every budget
(Section~\ref{sec:specificity}).
\item \textbf{A pre-registered negative.} The hypothesis that concentration explains the pruning
cost is tested and unsupported; we report it as unresolved, not reversed
(Section~\ref{sec:concentration}).
\item \textbf{Accuracy--perplexity decoupling.} The damage is visible in accuracy but not
perplexity; at 7B the two endpoints disagree in sign (Section~\ref{sec:target}).
\end{itemize}

\section{Related Work}
\paragraph{Reasoning-aware pruning.} Reconstruction-based \citep{frantar2023sparsegpt} and
activation-magnitude \citep{sun2024wanda} pruning are reasoning-blind. Recent methods inject
self-generated CoT into calibration \citep{lucas2025rac} or score components by causal
contribution and protect the important ones \citep{sheth2026cap}. Structured pruning of whole
units, including attention heads, has a long history \citep{michel2019sixteen,voita2019analyzing,ma2023llmpruner}.
We do not propose a new mask; we audit the reliability of the importance signal these methods
consume, which is complementary and, we argue, prior.

\paragraph{Causal attribution and circuit localization.} Our estimators follow the
activation-patching tradition \citep{conmy2023acdc} and its gradient approximation, attribution
patching \citep{syed2023attribution}, whose known failure modes at fine granularity our
granularity result quantifies in the reasoning setting. Interpretability work reports that
specific heads carry disproportionate functional load \citep{michel2019sixteen,voita2019analyzing};
we ask whether such localization transfers into a reliable pruning prior, and find specific
limits.

\section{Setup}\label{sec:setup}
\paragraph{Models.} We use a $2\times2$ grid over distillation and scale: two reasoning-distilled
models, DeepSeek-R1-Distill-Qwen-1.5B (D1) and -7B (D2) \citep{deepseek2025r1}, and two
size/family-matched non-distilled instruct models, Qwen2.5-1.5B-Instruct (N1) and
Qwen2.5-7B-Instruct (N2) \citep{qwen2024qwen25}. D1/D2 are distilled from the Qwen2.5 family,
making N1/N2 near-matched controls.

\paragraph{Scoring targets.} We measure the importance of a component by how much removing it
raises the model's loss on a chosen set of tokens. Concretely, head importance is the increase in
negative log-likelihood (NLL) on that token set when a query head's output is mean-ablated, that
is, replaced by its average over the calibration set (a standard way to delete a component's
contribution while keeping activation statistics in range). The \emph{scoring target} is precisely
this set of tokens: the choice of which tokens to score is the free decision that attribution
methods must make, and it is our object of study. We compare three targets: \ta{} = final-answer tokens with the CoT present (the naive
choice); \tc{} = the CoT reasoning tokens (the corrected choice); and \tr{} = answer tokens with
the CoT removed, a control isolating whether the pathology is answer tokens \emph{per se} or
answer-after-CoT.

\paragraph{Pruning and evaluation.} We prune at the granularity of whole query heads for three
reasons. First, the attribution-guided methods we audit operate on heads and comparable coarse
units, so this is where a reasoning-importance ranking actually drives a pruning mask. Second, a
head is structured hardware-friendly compute (an entire attention sub-computation), unlike an
individual weight. Third, our own granularity result (Section~\ref{sec:granularity}) shows that
importance is not reliably measurable at the level of single neurons, so the head is close to the
finest unit at which the ranking is trustworthy. Concretely, we prune a query head by zeroing the
corresponding rows of the query projection $W_Q$ and columns of the output projection $W_O$; under
grouped-query attention the key and value projections are shared across heads and we leave them
untouched (we assert a maximum $\|W_K\|,\|W_V\|$ norm change of exactly $0.0$ for every model). We report
accuracy (exact match, permissive extractor accepting \verb|\boxed{}|, ``the answer is $X$'', or
the final number), CoT-token NLL, and format-success rate on $\geq\!500$ held-out problems from
GSM8K, a grade-school math word-problem benchmark \citep{cobbe2021gsm8k}; calibration is drawn from the training split, so held-out evaluation is
disjoint by construction. To compare models at matched operating points we sweep the pruned-head
fraction and use each model's $f_{1/2}$, the fraction at which accuracy reaches half of dense,
requiring that the swept curve contains a point in the $25$--$50\%$-of-dense band (all
four models pass; no gap is computed on a floored curve). Calibration uses 128 correctly-answered
CoT trajectories per seed; each seed resamples calibration, recomputes importance, re-prunes, and
re-evaluates. We run \textbf{five seeds at 1.5B and three at 7B}; all intervals are hierarchical
bootstraps resampling both seeds and test items, and we report seed standard deviations rather
than item-only $p$-values on seed-averaged accuracy.

\paragraph{Compute and implementation.} Every experiment runs on a single NVIDIA H100 (80GB); no result requires multi-GPU execution,
and the full set of jobs totals roughly $75$ GPU-hours. Models are loaded in bfloat16 and decoded
greedily with a $1024$-token budget, so the only stochasticity in our pipeline is the calibration
draw. We use PyTorch $2.13$ and Transformers $5.13$; exact versions ship with the code release.

\paragraph{Operating points and the accuracy cliff.} Reasoning accuracy under head pruning
does not decay smoothly: it holds near dense and then falls sharply, which makes any single
sparsity level a poor basis for comparison and can floor a measurement entirely. We therefore
locate each model's operating point empirically (Table~\ref{tab:ops}) and require, as a
pre-registered gate, that the swept curve contain a point between $25\%$ and $50\%$ of dense
accuracy; only then is a gap computed on that curve. For the 1.5B models the cliff is steep
enough that a coarse sweep skipped the band and the gate correctly rejected it; refining to
single-head steps recovered a genuine slope. All four models pass with accuracy as the primary
endpoint, so no result in this paper is read off a floored curve. Reported variance is over
calibration resampling, which we find dominates: it is the seed standard deviation, not test-item
noise, that sets the width of our intervals.

\begin{table}[t]
\centering\small
\caption{Per-model operating points from the locating sweep (150 held-out items, one seed).
$f_{1/2}$ is the pruned-head count at which accuracy reaches half of dense. All four pass the
range-check gate, so accuracy is the primary endpoint throughout.}
\label{tab:ops}
\begin{tabular}{lccccc}
\toprule
Model & Total heads & Dense acc. & $f_{1/2}$ (heads) & $f_{1/2}$ (\%) & Range-check \\
\midrule
D1 & 336 & 62.7\% & 40  & 11.9\% & pass \\
N1 & 336 & 70.7\% & 44  & 13.1\% & pass \\
D2 & 784 & 65.3\% & 118 & 15.1\% & pass \\
N2 & 784 & 93.3\% & 118 & 15.1\% & pass \\
\bottomrule
\end{tabular}
\end{table}

\section{Answer-Token Attribution Saturates Universally}\label{sec:saturation}
The default scoring target is degenerate in every model we test. With the CoT in context, the
answer-token NLL is $9\times10^{-5}$ to $3.6\times10^{-3}$ (Table~\ref{tab:sat}): a CoT ends by
stating its answer, so re-predicting that answer is near-trivial and the target carries almost no
signal. What little signal remains is highly concentrated: in the distilled models one head out of
hundreds holds $\geq\!0.96$ of the total importance mass.

What causes this saturation is \emph{answer-in-CoT leakage}: a chain of thought typically states
the final answer within the reasoning text before the designated answer span, so once the CoT is in
context the answer tokens are nearly deterministic and carry little residual loss. We can measure
the strength of this leakage as the rise in answer-token NLL when the CoT is removed from context
(\tr{}$-$\ta): a large rise means the answer was hard to predict without the CoT and easy with it.
This leakage is universal rather than distillation-specific. The rise is large in all four models
($1.24$--$1.65$) and is in fact \emph{larger} in the non-distilled model N1 ($1.54$) than in the
distilled model D2 ($1.46$), so it is a property of step-by-step-then-answer generation itself, not
of distillation.
What does track distillation is how concentrated the resulting saturated signal becomes
(Table~\ref{tab:sat}, last column), which we examine in Section~\ref{sec:concentration}. Across
all models \ta{} ranks heads nearly orthogonally to \tr{} (Spearman $-0.008$ to $0.119$) and only
weakly agrees with \tc{} ($0.120$--$0.347$), so \ta{} is anomalous relative to \emph{both} other
targets. The pathology is the answer-after-CoT copy effect, not answer tokens broadly.

\begin{table}[t]
\centering\small
\caption{Answer-token attribution (\ta) saturates in all four models. Leakage (\tr$-$\ta) is
universal and is not larger in the distilled models; concentration of the saturated signal is.}
\label{tab:sat}
\setlength{\tabcolsep}{4pt}
\begin{tabular}{lccccc}
\toprule
Model & Distilled & \ta{} NLL & Leakage (\tr$-$\ta) & \ta{} Gini & \ta{} top-1 mass \\
\midrule
D1 (R1-Distill-Qwen-1.5B) & Y & 0.00065 & 1.651 & 0.987 & 0.969 \\
N1 (Qwen2.5-1.5B-Instruct) & N & 0.00020 & 1.540 & 0.916 & 0.260 \\
D2 (R1-Distill-Qwen-7B) & Y & 0.00009 & 1.456 & 0.997 & 0.996 \\
N2 (Qwen2.5-7B-Instruct) & N & 0.00362 & 1.244 & 0.843 & 0.598 \\
\bottomrule
\end{tabular}
\end{table}

\section{The Saturating Target Misguides Pruning}\label{sec:target}
\begin{figure}[t]
\centering
\includegraphics[width=\textwidth]{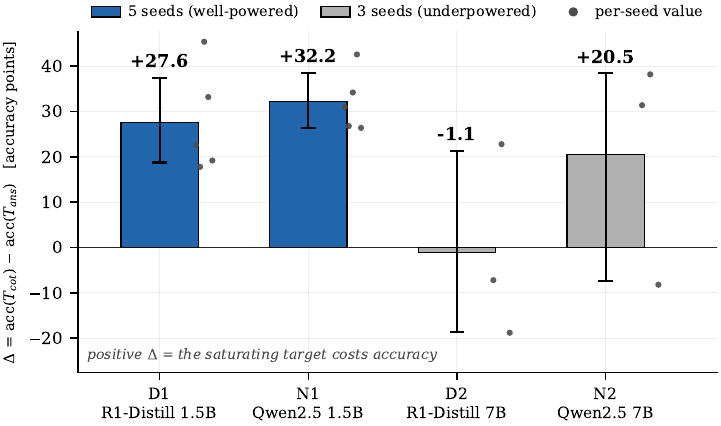}
\caption{The saturating target's pruning cost. Bars are
$\Delta=\mathrm{acc}(\tc)-\mathrm{acc}(\ta)$ at each model's $f_{1/2}$, with hierarchical-bootstrap
95\% intervals and per-seed values overlaid. The two five-seed 1.5B models (blue) show a large
cost that is positive in every seed; the three-seed 7B models (grey) are underpowered and we do
not draw conclusions from them.}
\label{fig:fig1}
\end{figure}

Saturation would be harmless if it did not change decisions; it does. Pruning by \ta{} rather than
\tc{} importance at each model's $f_{1/2}$ costs $27.6$ accuracy points on D1 (95\% CI
$[+18.8,+37.3]$, $p<0.001$) and $32.2$ on N1 (CI $[+26.4,+38.6]$, $p<0.001$), positive in all five
seeds of both models (Figure~\ref{fig:fig1}, Table~\ref{tab:prune}). The mechanism is visible in
the masks: because the saturated ranking assigns almost all mass to one head, it protects that head
and orders the rest near-arbitrarily, so accuracy collapses to near-chance and format-success falls
to $56$--$64\%$. The corrected target holds format-success above $97\%$ at the same sparsity. The
\tr{} control reproduces the effect ($\Delta(\tc-\tr)=+27.6$ on D1, $+35.9$ on N1), so the benefit
comes from scoring reasoning tokens rather than from any peculiarity of the answer span.

The 7B results point the same way but are underpowered at three seeds: N2 shows $+20.5$ (CI
$[-7.4,+38.4]$, $p=0.07$) and D2 shows $-1.1$ (CI $[-18.6,+21.2]$, $p=0.79$) with per-seed values
that flip sign ($-7.2$, $+22.8$, $-18.8$). We therefore treat D2 as \emph{unresolved}, not as a
null: three seeds cannot resolve an effect of this variance. The robust, well-powered conclusion
is at 1.5B: the naive answer-token target imposes a large reasoning-accuracy cost, and importance
should be scored on reasoning tokens.

\paragraph{Accuracy--perplexity decoupling.} The damage is not faithfully reflected in perplexity.
On D1, \ta-guided pruning inflates CoT-NLL by $+3.18$ alongside the accuracy collapse; but on D2
and N2 the NLL gap is \emph{negative} ($-0.20$, $-0.35$) even where the accuracy gap is positive.
The two endpoints disagree in sign at 7B. Perplexity-only evaluation of pruned reasoners can
therefore understate or misrank the damage, a caution that applies to any compression work
reporting perplexity alone.

\begin{figure}[t]
\centering
\includegraphics[width=\textwidth]{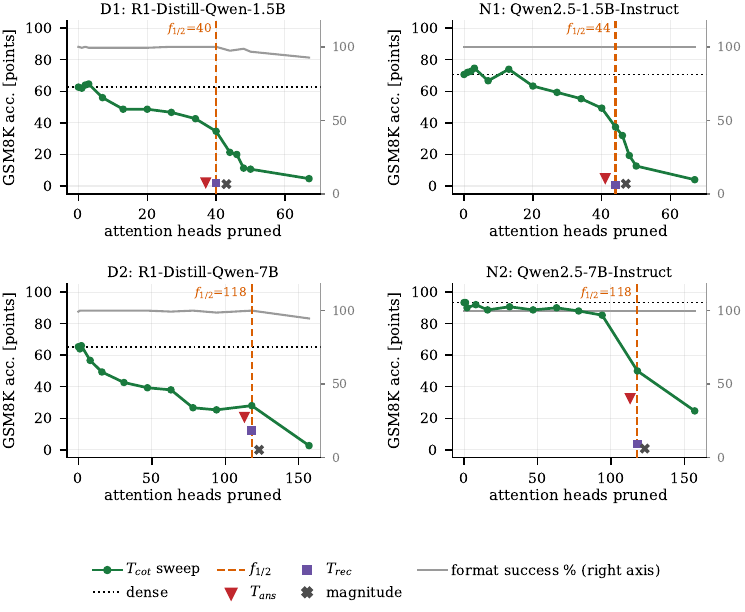}
\caption{Accuracy versus pruned heads under the corrected target \tc{} (green), with \ta{}
($\blacktriangledown$), the \tr{} control ($\blacksquare$) and a target-agnostic weight-magnitude
baseline ($\times$) evaluated at each model's $f_{1/2}$ (orange dashed line); markers are offset
horizontally so they remain distinguishable where they coincide near the floor. The grey line and
right-hand axis give format-success rate, which stays high along the \tc{} sweep, so the decline
is loss of reasoning rather than loss of output structure. The magnitude baseline sits at or near the
floor at $f_{1/2}$ in all four models, so the \tc{} advantage is not recoverable from a
reasoning-blind score.}
\label{fig:fig2}
\end{figure}

\paragraph{The gain is not recoverable from a reasoning-blind score.} A natural objection is that the
corrected target merely recovers what a standard magnitude criterion would give. It does not. We
include a target-agnostic weight-magnitude baseline (the sum of $W_Q$-row and $W_O$-column norms
per head) at every operating point (Figure~\ref{fig:fig2}, $\times$): at $f_{1/2}$ it sits at or
near the floor in all four models, for example $1.2\%$ and $0.8\%$ accuracy on D1 at $40$ and
$44$ heads. The signal that \tc{} supplies is therefore not a proxy for weight magnitude. We also emphasise that we do not propose a mask: our object of study is the importance
signal that a mask consumes, and our recommendations apply to any method that consumes one.

\begin{table}[t]
\centering\small
\caption{Pruning by the saturating target (\ta) versus the corrected target (\tc) at each model's
own $f_{1/2}$ (in heads). $\Delta>0$ means the saturating target costs accuracy. The seed standard
deviation makes the power difference explicit: it is $5.9$--$10.4$ points on the five-seed 1.5B
models but $17.5$--$20.5$ on the three-seed 7B models, which is why we label D2 unresolved rather
than null. Intervals are hierarchical bootstraps over seeds and test items.}
\label{tab:prune}
\begin{tabular}{lccccccc}
\toprule
Model & Seeds & Dense & $f_{1/2}$ & $\Delta$ (points) & seed s.d. & 95\% CI & $p$ \\
\midrule
D1 & 5 & 63.0\% & 40  & $+27.6$ & $10.4$ & $[+18.8,+37.3]$ & $<0.001$ \\
N1 & 5 & 74.0\% & 44  & $+32.2$ & $\phantom{0}5.9$ & $[+26.4,+38.6]$ & $<0.001$ \\
D2 & 3 & 71.2\% & 118 & $-1.1$ \emph{(unres.)} & $17.5$ & $[-18.6,+21.2]$ & $0.79$ \\
N2 & 3 & 91.2\% & 118 & $+20.5$ & $20.5$ & $[-7.4,+38.4]$ & $0.07$ \\
\bottomrule
\end{tabular}
\end{table}

\section{Concentration Does Not Explain the Pruning Cost}\label{sec:concentration}
Sections~\ref{sec:saturation} and~\ref{sec:target} established two facts about the saturating
target: its importance mass concentrates onto a single head, and pruning with it is costly. Before
the pre-registration we conjectured that the first explains the second, and we report the test of
that conjecture here.

Concentration tracks distillation sharply (Table~\ref{tab:sat}): the \ta{} top-1 head holds
$0.969$/$0.996$ of the mass in D1/D2 versus $0.260$/$0.598$ in N1/N2. A robustness diagnostic
indicates this is structural rather than an artifact of near-zero scores: in the distilled models
the dominant head is identical across all five calibration resamples and decisive over the
runner-up ($113\times$ for D1, $6539\times$ for D2), and it coincides with the top head under the
corrected target \tc; in non-distilled models it is unstable across resamples (N1, two distinct
winners) or far less decisive (N2, $16\times$). D1 and D2 concentrate onto the same head position
across a $4.6\times$ scale gap.

We pre-registered the prediction that this concentration explains the pruning cost: if \ta{}
collapses onto one head, \ta-guided pruning protects one head and prunes the rest near-randomly,
so more concentrated models should show larger gaps. \textbf{The prediction is not supported.}
Across the four models the rank correlation between concentration and $\Delta$ is $-0.80$, if
anything inverse: the most concentrated model (D2, $0.996$) shows the smallest, non-significant
gap and the least concentrated (N1, $0.260$) the largest. Two facts prevent us from claiming a
genuine inverse effect. The correlation rests on four points, two of which (D2, N2) are
noise-dominated at three seeds; and concentration is confounded with scale in this grid, since both
high-concentration models are the 7B ones, so the trend may reflect ``7B is more robust to head
pruning'' rather than anything about concentration. We therefore report the concentration
mechanism as \textbf{tested and unresolved}, and defer a de-confounded test (concentration varied
within a fixed scale, with a powered 7B arm) to future work.

\begin{figure}[t]
\centering
\includegraphics[width=0.48\textwidth]{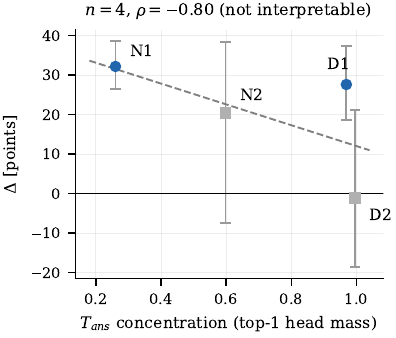}
\caption{The pre-registered concentration mechanism, shown for completeness and \emph{not} as a
result. Each point is one model ($n=4$); grey squares are the underpowered three-seed 7B models.
With four points, two of them noise-dominated, and concentration confounded with scale, this
panel cannot support either a positive or an inverse mechanism claim.}
\label{fig:fig4}
\end{figure}

We report this negative in full
because it is the honest outcome of a pre-registered prediction; we draw no mechanism claim from
it, and none of our recommendations depend on it.

\section{Reasoning Importance Is Not Finely Localizable}\label{sec:granularity}
\begin{figure}[t]
\centering
\includegraphics[width=\textwidth]{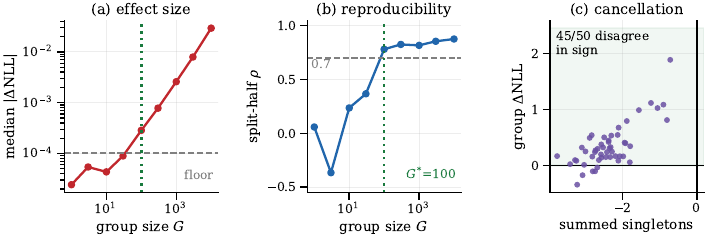}
\caption{Reasoning importance is not localizable below $\sim\!100$ neurons. (a) Median
$|\Delta$NLL$|$ from ablating $G$ neurons grows with $G$; single neurons sit at the measurement
floor (dashed line, estimated from a rounding-scale sham perturbation). (b) Split-half Spearman
between independent calibration halves crosses the $0.7$ bar only at $G^*\!=\!100$ (green dotted
line in both panels). (c) Cancellation: each point is one 100-neuron group, plotting the group's
measured $\Delta$NLL against the signed sum of its members' individual effects (both axes in units
of $10^{-3}$). The summed singletons are negative for all 50 groups while the group effect is
positive for 45 of them, so single-unit effects do not compose.}
\label{fig:fig3}
\end{figure}

The target choice was the first of the three attribution decisions from Section~\ref{sec:intro};
we now turn to the second, the granularity at which importance is scored. Here the fragility is not
that a plausible choice misfires but that the finest choice is ill-posed: causal reasoning-importance
is not measurable at single-unit granularity. Ablating groups of $G$
MLP neurons (individual feed-forward hidden units, the post-SwiGLU activations feeding
\verb|down_proj|; $8{,}960$ per layer and $250{,}880$ in total for D1) and measuring the
effect on CoT-token NLL, single-neuron effects sit at the measurement floor
($2.4\times10^{-5}$ median, against a floor of $1.0\times10^{-4}$ estimated from a
rounding-scale sham perturbation) and their cross-split ranking is noise. Here \emph{split-half
Spearman} denotes the rank agreement between importance scores computed on two disjoint halves of
the calibration set, where $1$ is perfect agreement and $0$ is chance; for single neurons it is
$0.059$ at $G{=}1$ and $-0.366$ at $G{=}3$. Reproducible importance emerges only near
$G\!\approx\!100$: split-half Spearman rises from $0.367$ at $G{=}30$ to $0.781$ at $G{=}100$ and
stabilises above $0.8$ thereafter (Figure~\ref{fig:fig3}b). Two independent criteria, the
reproducibility onset and the numerical-floor crossing, both place $G^*$ at $100$.

The mechanism is cancellation, not mere redundancy. Across $50$ groups of $100$ neurons, the
group's ablation effect is positive in $90\%$ of cases while the signed sum of its members'
individual effects is negative in \emph{all} $50$, a sign disagreement of $0.90$
(Figure~\ref{fig:fig3}c), with the group effect a median $0.105\times$ the magnitude of the
summed singletons. Individual neuron effects do not compose; they cancel. Single-neuron causal
attribution, sometimes proposed as a finer-grained refinement of head-level methods, is therefore
numerically ill-posed at this scale, independent of estimator: the quantity being estimated is
itself below the floor.

\paragraph{Reconciling with head-level sensitivity.} This may appear to contradict the finding
that individual \emph{heads} matter a great deal (Section~\ref{sec:target}, and the appendix
probe). It does not. A head is a large routing object, $128$ dimensions writing to the residual
stream through $W_O$, whereas an MLP neuron is one of $8{,}960$ per layer. Redundancy and
cancellation operate at the sub-unit scale; heads are already coarse enough to be individually
measurable. Our result is a statement about \emph{where the measurement floor sits}, and it places
that floor between the neuron and the head.

\section{Reasoning-Specific Importance Is Not Separable}\label{sec:specificity}
The third attribution decision is whether reasoning-specific components can be isolated from
generically important ones, as circuit-localization work suggests and as a protection scheme might
exploit. Contrastively isolated ``reasoning-specific'' importance, defined as reasoning contribution
minus a matched non-reasoning baseline, is not separable from raw importance for masking. At head level the reasoning-specific contrast collapses to its own
split-half noise floor: the divergence between the specific ranking and raw importance is
$1.00$, $1.08$ and $0.97$ times that floor at $k=5$, $10$ and $20\%$. That is, the specific
ranking disagrees with raw importance no more than it disagrees with itself across data splits.
Introducing a length- and token-type-matched control (shuffled-CoT, itself highly reproducible at
split-half Spearman $0.983$) shrinks the apparent divergence further: the rank correlation with
raw importance rises to $0.509$ and the top-$k$ overlap ratio ($0.71$--$0.79$) fails a pre-set
separation bar of $0.7$.

At neuron level the decision-relevant test is direct: ablating the top reasoning-specific
group is \emph{less} damaging to reasoning than ablating the top raw-importance group at every
budget ($\Delta$NLL $0.42$ vs $0.70$ at $1\%$, $0.98$ vs $1.31$ at $5\%$, $1.47$ vs $1.68$ at
$10\%$; random groups $0.007$--$0.085$). Protecting the ``specific'' set is thus worse than
protecting the generically important set. Reasoning-specific components, insofar as they can be
measured at all, are largely the generically important components.

\section{Discussion and Recommendations}
The three fragilities yield concrete guidance for attributing reasoning-importance in CoT
models, for pruning and equally for analysis or data selection.
\begin{itemize}\itemsep2pt
\item \textbf{Score reasoning tokens, not answer tokens.} Answer-token targets saturate in every
CoT model tested and, for pruning, cost $27$--$32$ accuracy points at 1.5B. Score the CoT span.
\item \textbf{Do not attribute at single-unit granularity.} Neuron-level causal importance cancels
and is irreproducible; use groups of $\gtrsim\!100$ units, or coarser objects such as heads.
\item \textbf{Do not rely on reasoning-specific localization.} Specific and generic importance are
not separable for masking; contrastive protection does not beat raw importance.
\item \textbf{Evaluate with accuracy, not perplexity.} The two decouple under reasoning pruning
and disagree in sign at 7B.
\end{itemize}
Each is inexpensive to apply, and they complement rather than replace existing reasoning-aware
pruning methods.

\paragraph{Scope of the claims.} We are explicit about what each finding is powered to support.
The universal saturation of \ta{} (Section~\ref{sec:saturation}) is measured in \emph{all four}
models and is the broadest claim. The pruning cost (Section~\ref{sec:target}) is established at
1.5B with five seeds on GSM8K, and we deliberately do not generalise it to 7B or to other tasks;
the 7B arm is reported so that a reader can see its variance rather than omitted because it is
inconvenient. The granularity and specificity results (Sections~\ref{sec:granularity}
and~\ref{sec:specificity}) are measured on one model family and we present them as such. The
concentration mechanism is unresolved. We would rather state a narrow claim that holds than a
broad one that does not.

\section{Limitations}
Our well-powered evidence is at 1.5B on GSM8K with five seeds; the 7B arm has three seeds, does
not resolve the concentration hypothesis, and leaves D2's gap indeterminate rather than null.
Concentration is confounded with scale in our $2\times2$ grid, so Section~\ref{sec:concentration}
cannot separate the two. We study attention-head pruning; head-level protection is in tension with
the 2:4 semi-structured sparsity formats (two nonzero weights per group of four, which GPUs accelerate) that deliver kernel speedups, and we do not report end-to-end
latency, so our claims are about the \emph{signal}, not about achieved compression. Our estimators
are mean-ablation for heads and attribution patching for neurons; the latter did not pass a
per-neuron validation gate (rank-stratified Spearman $0.65$ against true ablation), which is why we
report neuron results only at the collective, group level, and which is itself consistent with the
granularity finding. Task coverage is math reasoning; extending to non-math CoT is future work.
The single-head probe in Appendix~\ref{app:head} is preliminary and uncontrolled, and we draw no
claim from it.

\section{Conclusion}
Attribution-guided pruning of reasoning models rests on an importance signal that is fragile to
measure. The natural answer-token target saturates in every CoT model and, used for pruning,
collapses reasoning accuracy while barely moving perplexity; causal importance is not localizable
below $\sim\!100$ neurons because single-unit effects cancel; and reasoning-specific importance is
not separable from generic importance. A pre-registered concentration mechanism is tested and
unresolved. The practical consequence is a set of measurement rules that make reasoning-importance attribution
reliable, offered as a complement to the reasoning-aware pruning literature.

\bibliographystyle{colm2026_conference}
\bibliography{colm2026_conference}

@inproceedings{han2015learning,
  title={Learning both Weights and Connections for Efficient Neural Networks},
  author={Han, Song and Pool, Jeff and Tran, John and Dally, William J.},
  booktitle={Advances in Neural Information Processing Systems (NeurIPS)},
  year={2015},
  note={arXiv:1506.02626}
}

@inproceedings{frantar2023sparsegpt,
  title={{SparseGPT}: Massive Language Models Can Be Accurately Pruned in One-Shot},
  author={Frantar, Elias and Alistarh, Dan},
  booktitle={International Conference on Machine Learning (ICML)},
  year={2023},
  note={arXiv:2301.00774}
}

@inproceedings{sun2024wanda,
  title={A Simple and Effective Pruning Approach for Large Language Models},
  author={Sun, Mingjie and Liu, Zhuang and Bair, Anna and Kolter, J. Zico},
  booktitle={International Conference on Learning Representations (ICLR)},
  year={2024},
  note={arXiv:2306.11695}
}

@inproceedings{michel2019sixteen,
  title={Are Sixteen Heads Really Better than One?},
  author={Michel, Paul and Levy, Omer and Neubig, Graham},
  booktitle={Advances in Neural Information Processing Systems (NeurIPS)},
  year={2019},
  note={arXiv:1905.10650}
}

@inproceedings{voita2019analyzing,
  title={Analyzing Multi-Head Self-Attention: Specialized Heads Do the Heavy Lifting, the Rest Can Be Pruned},
  author={Voita, Elena and Talbot, David and Moiseev, Fedor and Sennrich, Rico and Titov, Ivan},
  booktitle={Annual Meeting of the Association for Computational Linguistics (ACL)},
  year={2019},
  note={arXiv:1905.09418}
}

@inproceedings{ma2023llmpruner,
  title={{LLM-Pruner}: On the Structural Pruning of Large Language Models},
  author={Ma, Xinyin and Fang, Gongfan and Wang, Xinchao},
  booktitle={Advances in Neural Information Processing Systems (NeurIPS)},
  year={2023},
  note={arXiv:2305.11627}
}

@article{lucas2025rac,
  title={Reasoning Models Can be Accurately Pruned Via Chain-of-Thought Reconstruction},
  author={Lucas, Ryan and Behdin, Kayhan and Wang, Zhipeng and Song, Qingquan and Tang, Shao and Mazumder, Rahul},
  journal={arXiv preprint arXiv:2509.12464},
  year={2025}
}

@article{sheth2026cap,
  title={Pruning via Causal Attribution Preserves Reasoning Performance in Large Language Models},
  author={Sheth, Amogh and Assefa, Biruk and Huang, Yi Wen and Lin, Andrew and Ge, Yuhao},
  journal={arXiv preprint arXiv:2606.19350},
  year={2026}
}

@inproceedings{conmy2023acdc,
  title={Towards Automated Circuit Discovery for Mechanistic Interpretability},
  author={Conmy, Arthur and Mavor-Parker, Augustine N. and Lynch, Aengus and Heimersheim, Stefan and Garriga-Alonso, Adri{\`a}},
  booktitle={Advances in Neural Information Processing Systems (NeurIPS)},
  year={2023},
  note={arXiv:2304.14997}
}

@article{syed2023attribution,
  title={Attribution Patching Outperforms Automated Circuit Discovery},
  author={Syed, Aaquib and Rager, Can and Conmy, Arthur},
  journal={arXiv preprint arXiv:2310.10348},
  year={2023}
}

@article{cobbe2021gsm8k,
  title={Training Verifiers to Solve Math Word Problems},
  author={Cobbe, Karl and Kosaraju, Vineet and Bavarian, Mohammad and Chen, Mark and Jun, Heewoo and Kaiser, Lukasz and Plappert, Matthias and Tworek, Jerry and Hilton, Jacob and Nakano, Reiichiro and Hesse, Christopher and Schulman, John},
  journal={arXiv preprint arXiv:2110.14168},
  year={2021}
}

@inproceedings{wei2022chain,
  title={Chain-of-Thought Prompting Elicits Reasoning in Large Language Models},
  author={Wei, Jason and Wang, Xuezhi and Schuurmans, Dale and Bosma, Maarten and Ichter, Brian and Xia, Fei and Chi, Ed H. and Le, Quoc V. and Zhou, Denny},
  booktitle={Advances in Neural Information Processing Systems (NeurIPS)},
  year={2022},
  note={arXiv:2201.11903}
}

@article{deepseek2025r1,
  title={{DeepSeek-R1}: Incentivizing Reasoning Capability in LLMs via Reinforcement Learning},
  author={{DeepSeek-AI} and Guo, Daya and Yang, Dejian and Zhang, Haowei and others},
  journal={arXiv preprint arXiv:2501.12948},
  year={2025}
}

@article{qwen2024qwen25,
  title={{Qwen2.5} Technical Report},
  author={{Qwen Team}},
  journal={arXiv preprint arXiv:2412.15115},
  year={2024}
}

\appendix
\section{Preliminary single-head ablation (not controlled; no claim drawn)}\label{app:head}
As a preliminary probe of the concentrated head's causal role, ablating that single head alone
reduced GSM8K accuracy from $60.4\%$ to $2.6\%$ on D1 and from $68.8\%$ to $1.4\%$ on D2 (its own
dense decode; $500$ held-out items). Ablating a random early-layer head instead cost $-0.6$ points
on average on D1 (three draws: $-0.4$, $+3.2$, $-4.6$).

\textbf{We draw no conclusion from this.} The random baseline is three draws, not a distribution;
on D2 the three random early-layer draws were $+1.2$, $-1.8$ and $+45.4$ points of damage (mean
$+14.9$), i.e.\ one randomly chosen early-layer head was itself worth $45$ points, so at 7B the
target head is not cleanly separable from other early-layer heads. We have also not disentangled
reasoning degradation from generation/format collapse. A controlled version, ablating the target
head against a distribution of 10 to 20 random heads per layer band, with format-success reported
and a matched-capacity control, is required before any single-head causal claim, and is left to
future work.

\section{Reproducibility}
We report all seeds, calibration draws, held-out test identifiers and per-job artifact hashes with
the code release.
\ifcolmsubmission
The code, data-processing scripts, and figure-generation scripts will be released publicly; the URL
is withheld here to preserve anonymity and will be added in the camera-ready version.
\else
The code, data-processing scripts, and figure-generation scripts are available at
\url{https://github.com/himans-iitk/Reasoning_in_Pruning}.
\fi Pruning is deterministic given an importance vector; reported variance is over
calibration resampling (five seeds at 1.5B, three at 7B). Intervals are hierarchical bootstraps
over seeds and test items. Key-value projections are verified unchanged under head pruning
(maximum norm delta $0.0$ in all four models).

\end{document}